\documentclass[lettersize,journal]{IEEEtran}
\usepackage{amsmath,amsfonts}
\usepackage{algorithmic}
\usepackage{algorithm}
\usepackage{array}
\usepackage[caption=false,font=normalsize,labelfont=sf,textfont=sf]{subfig}
\usepackage{textcomp}
\usepackage{stfloats}
\usepackage{url}
\usepackage{verbatim}
\usepackage{graphicx}
\usepackage{cite}
\usepackage{hyperref}
\usepackage{multirow}
\usepackage{dsfont} 

\usepackage{tikz,xcolor}

\hypersetup{hidelinks,
	colorlinks=true,
	allcolors=black,
	pdfstartview=Fit,
	breaklinks=true}
	
\definecolor{lime}{HTML}{A6CE39}
\DeclareRobustCommand{\orcidicon}{
\begin{tikzpicture}
\draw[lime, fill=lime] (0,0)
circle[radius=0.16]
node[white]{{\fontfamily{qag}\selectfont \tiny \.{I}D}};
\end{tikzpicture}
\hspace{-2mm}
}
\foreach \x in {A, ..., Z}{%
\expandafter\xdef\csname orcid\x\endcsname{\noexpand\href{https://orcid.org/\csname orcidauthor\x\endcsname}{\noexpand\orcidicon}}
}

\begin{document}

\title{AIM: Adaptive Intra-Network Modulation for Balanced Multimodal Learning}
\author{Shu Shen\hspace{-1.5mm}\orcidB{},~\IEEEmembership{Student Member,~IEEE,} C. L. Philip Chen\hspace{-1.5mm}\orcidD{},~\IEEEmembership{Life Fellow,~IEEE, } and Tong Zhang\IEEEauthorrefmark{1}\thanks{* Corresponding author: Tong Zhang.}\hspace{-1.5mm}\orcidA{},~\IEEEmembership{Senior Member,~IEEE}
\thanks{This work was funded in part by the National Natural Science Foundation of China grant under number 62222603, in part by the STI2030-Major Projects grant from the Ministry of Science and Technology of the People’s Republic of China under number 2021ZD0200700, in part by the Key-Area Research and Development Program of Guangdong Province under number 2023B0303030001, in part by the Program for Guangdong Introducing Innovative and Entrepreneurial Teams (2019ZT08X214), and in part by the Science and Technology Program of Guangzhou under number 2024A04J6310.}
\thanks{The authors are with the Guangdong Provincial Key Laboratory of Computational AI Models and Cognitive Intelligence, the School of Computer Science and Engineering, South China University of Technology, Guangzhou 510006, China, and is with the Pazhou Lab, Guangzhou 510335, China, and is with Engineering Research Center of the Ministry of Education on Health Intelligent Perception and Paralleled Digital-Human, Guangzhou, China. (e-mail: tony@scut.edu.cn).}}


\markboth{Journal of \LaTeX\ Class Files,~Vol.~14, No.~8, October~2025}%
{Shell \MakeLowercase{\textit{et al.}}: A Sample Article Using IEEEtran.cls for IEEE Journals}


\maketitle

\begin{abstract}
Multimodal learning has substantially improved machine learning performance, yet it still faces the problem of modality imbalance, which hinders the adequate leverage of different modalities. Existing methods typically adopt network-level modulation mechanisms for balanced multimodal learning, stimulating the entire network of weak modalities while mitigating the dominant ones. However, these methods generally overlook the differences in optimization pace across parameters and layers within the network. Consequently, slow-optimizing network components in dominant modalities are over-suppressed, whereas those in weak modalities are under-stimulated, hindering their full learning potential. To address this issue, we propose Adaptive Intra-Network Modulation (AIM), which introduces a training framework that integrates parameter-adaptive modulation (PAM) and depth-adaptive modulation (DAM). PAM adaptively decouples and stimulates under-optimized parameters of each modality while balancing their learning, guided by their performance at each depth. DAM then adaptively strengthens PAM across depths according to their imbalance levels. In this way, the multimodal learning is balanced with the learning potential of each modality network fully exploited. Additionally, Depth-Adaptive Prototype is proposed to facilitate intra-network modulation. It accounts for differences in optimization capacity across network layers, providing an effective strategy for evaluating the performance of different components within the network. Experimental results show that AIM outperforms state-of-the-art balanced multimodal learning methods across multiple benchmarks and is effective with various fusion strategies, optimizers, and backbones.

\end{abstract}


\section{Introduction} \label{sec:intro}
\IEEEPARstart{H}{umans} perceive and understand the surrounding world by integrating multiple senses, such as vision, hearing, and touch \cite{gazzaniga2006cognitive,herreras2010cognitive}. Inspired by this, multimodal learning \cite{dou2022empirical,shankar2018review,10339893,liang2022foundations,baltruvsaitis2018multimodal,10814984,9681296,10171388}, which leverages and fuses data from diverse sensors, has attracted considerable attention and achieved substantial progress. In recent years, it has significantly improved the performance of machine learning in various applications, including action recognition \cite{kazakos2019epic,gao2020listen}, emotion recognition \cite{hazarika2020misa,sun2022cubemlp}, and audio-visual speech recognition \cite{potamianos2004audio,8387512}, and other multimedia tasks \cite{10445009,1468162}.

\begin{table}[t]
\caption{Performance comparison of different modalities among unimodal-only models, joint-training multimodal baseline, several state-of-the-art balanced multimodal learning methods (OGM-GE, PMR, and MLA), and our proposed AIM. * represents the dominant modality with higher performance.}
\begin{center}
\begin{tabular}{c|c|c|c|c}
\hline
\multirow{2}{*}{Method} & \multicolumn{2}{c|}{CREMA-D}        & \multicolumn{2}{c}{Kinetics-Sounds}  \\ 
                        & \multicolumn{1}{c}{Audio} & Visual* & \multicolumn{1}{c}{Audio*}  & Visual \\ \hline
Audio-only model               & \multicolumn{1}{c}{63.17} & -- & \multicolumn{1}{c}{60.44} & -- \\ 
Visual-only model               & \multicolumn{1}{c}{--} & 68.15 & \multicolumn{1}{c}{--} & 37.78 \\ \hline
Joint-training                  & \multicolumn{1}{c}{58.72} & 66.59 & \multicolumn{1}{c}{58.61} & 30.42 \\ \hline
OGM-GE \cite{peng2022balanced} & \multicolumn{1}{c}{60.32} & 65.73 & \multicolumn{1}{c}{55.83} & 34.55 \\ 
PMR \cite{fan2023pmr} & \multicolumn{1}{c}{59.02} & 67.83 & \multicolumn{1}{c}{57.63} & 33.14 \\
MLA \cite{zhang2024multimodal} & \multicolumn{1}{c}{61.26} & 67.49 & \multicolumn{1}{c}{57.84} & 34.31 \\ \hline
Ours                  & \multicolumn{1}{c}{\underline{63.03}} & \textbf{69.21} & \multicolumn{1}{c}{\textbf{63.16}} & \underline{37.27} \\ \hline
\end{tabular}
\end{center}
\label{tab:winwin}
\end{table}

\begin{figure}[t]
\centering
\includegraphics[width=\columnwidth]{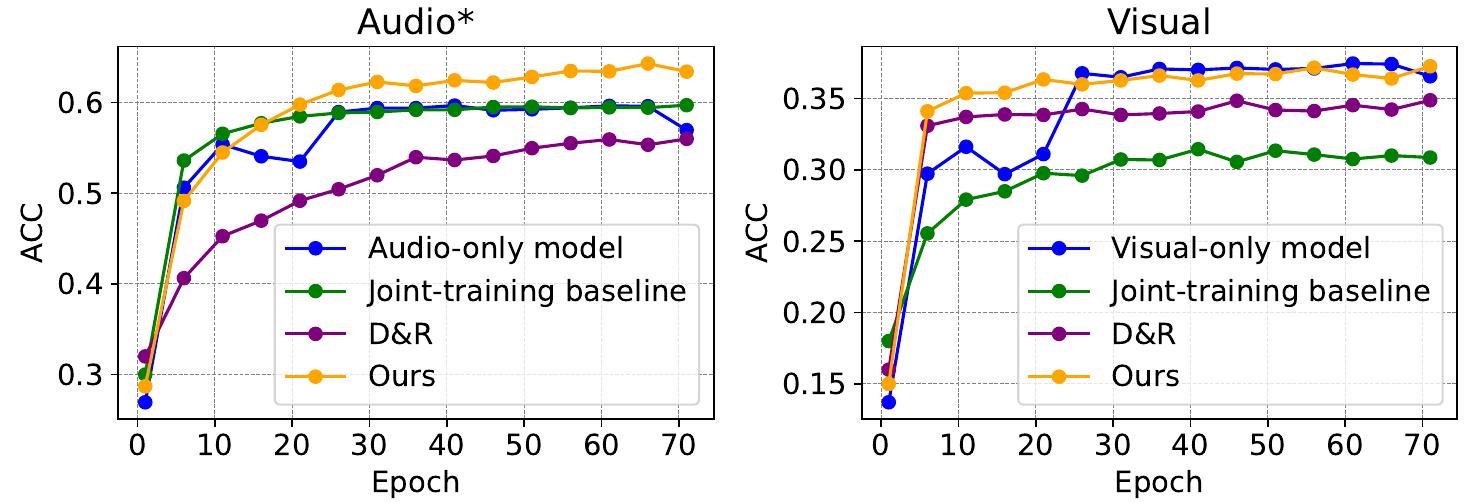} 
\caption{Accuracy variation of each modality while training on the Kinetics-Sounds dataset for unimodal-only models, the joint-training multimodal baseline, the state-of-the-art balanced multimodal learning method D\&R \cite{wei2024diagnosing}, and our proposed AIM. * represents the dominant modality with higher performance.}
\label{fig:winwin}
\end{figure}

Despite rapid progress, recent studies have highlighted that modality imbalance \cite{huang2022modality,peng2022balanced,wang2020makes,wei2024fly} severely affects the performance of multimodal learning. This issue stems from discrepancies in performance and convergence rates across modalities, leading to uncoordinated convergence under a unified joint training objective \cite{peng2022balanced,wei2024fly}. Consequently, weak modalities with poorer performance are suppressed by dominant ones with better performance during optimization, hindering their sufficient learning. To address this issue, various methods \cite{wang2020makes,peng2022balanced,wu2022characterizing,fan2023pmr,wei2024diagnosing,wei2024fly} have proposed modulation strategies for each modality, stimulating the learning of the entire weak modality networks or mitigating that of dominant ones. For example, Peng et al. \cite{peng2022balanced} and Wei et al. \cite{wei2024fly} scale the gradients of all parameters in each modality network by different factors. Fan et al. \cite{fan2023pmr} define a rebalance loss for each modality network, while Wei et al. \cite{wei2024diagnosing} reinitialize all parameters of each modality network with different strengths. 

However, we observe that these modulation methods fail to fully exploit the learning potential of each modality network and may even hinder dominant modalities when promoting weaker ones. As shown in Fig. \ref{fig:winwin} and Table \ref{tab:winwin}, the performance of each modality under these modulation methods is inferior to that of the corresponding unimodal-only networks. Moreover, compared with the joint-training baseline, although these methods improve the performance of weak modalities, the dominant modalities achieve comparable or even lower performance and exhibit slower convergence. We analyze this result and consider that the network-level design of these methods impedes the sufficient optimization of each modality network. As revealed by recent studies \cite{saratchandran2024activation,chen2023layer,liu2025optimization}, due to factors such as parameter initialization \cite{saratchandran2024activation} and gradient instability \cite{chen2023layer,liu2025optimization}, different parameters and layers within a deep network exhibit varying optimization paces during training. However, these modulation methods overlook this issue, applying the same degree of suppression or stimulation to all parameters and layers within dominant or weak modality networks. Therefore, in dominant modality networks, parameters or layers that optimize more slowly are over-suppressed, leading to performance decline after modulation. In weak modality networks, although overall suppression is alleviated, stimulation of slow-optimizing parameters or layers remains insufficient, resulting in performance far below that of the corresponding unimodal models. Thus, we pose a key question: \textbf{\textit{How to design an modulation mechanism to stimulate slow-optimizing parameters and layers in different modality networks, thereby fully exploiting their optimization potential and enhancing balanced multimodal learning?}}




This paper addresses the above question by proposing \underline{A}daptive \underline{I}ntra-Network \underline{M}odulation (AIM) for balanced multimodal learning. AIM introduces a training framework that incorporates \textbf{\textit{parameter-adaptive modulation (PAM)}} and \textbf{\textit{depth-adaptive modulation (DAM)}}. This paradigm enables sufficient learning of parameters and layers with slower optimization paces when balancing the learning of different modalities, fully realizing the optimization potential of each modality network. Furthermore, \textbf{\textit{depth-adaptive prototypes (DAP)}} is proposed for intra-network modulation. Its design accounts for the varying optimization capacities across network layers, providing an effective strategy for evaluating performance across different components within the network.

Specifically, during training, each unimodal network is first partitioned into an equal number of sequential network blocks. At each depth, the \textbf{\textit{parameter-adaptive modulation (PAM) }}of AIM first introduces a parameter decoupling module that extracts the low-performing, under-optimized parameters from each modality network block, forming new network blocks referred to as Auxiliary Blocks. Subsequently, the original network blocks and Auxiliary Blocks of all modalities at the same depth are jointly trained, with their participation degrees adjusted based on their performance evaluated via DAP. In particular, \textbf{for weak modalities}, the suppression they encounter is effectively mitigated by reducing the involvement of high-performance original network blocks from dominant modalities. In addition, the introduction of Auxiliary Blocks fully stimulates the learning of their parameters with slower optimization paces. \textbf{For dominant modalities}, although direct learning of their original network blocks is reduced, their optimization is nevertheless enhanced by focusing training on the under-optimized parameters via their Auxiliary Blocks. PAM also promotes the learning of layers with slower optimization pace in each modality network by explicitly considering their performance at each depth. Subsequently, the \textbf{\textit{depth-adaptive modulation (DAM)}} is applied to address the variable modality imbalance level arising from differences in optimization pace across depths. It evaluates performance discrepancies of modalities at each depth and accordingly strengthens PAM to depths with greater imbalance, thereby enhancing balanced multimodal learning.

The contributions of this paper can be summarized as:
\begin{itemize}
    \item This work focuses on a commonly overlooked issue in balanced multimodal learning, namely the differences in optimization pace across network layers and parameters. We accordingly propose Adaptive Intra-Network Modulation (AIM), which outperforms existing methods across multiple benchmarks.

    \item AIM introduces a training framework that incorporates parameter-adaptive and depth-adaptive modulation. This framework enables sufficient learning of components with slower optimization within each modality network, achieving balanced multimodal learning with each modality network’s learning potential fully exploited.

    \item We propose a performance evaluation method based on depth-adaptive prototypes for intra-network modulation. It fully accounts for the differences in optimization capacity across network layers and provides an effective approach for assessing the performance of different components within the network.

\end{itemize}

\section{Related Works}
Due to space limitations, this section only presents part of the related works. \textbf{More detailed discussions, including reviews of studies on \textit{Differences of Optimization Pace In Deep Neural Networks} and \textit{Prototypical Methods}, are provided in the Supplemental Materials.}

\subsection{Multimodal Learning}

With advances in sensor technologies and data processing, multimodal learning has rapidly developed and substantially improved machine learning performance across a wide range of tasks and applications \cite{ramachandram2017deep,baltruvsaitis2018multimodal,zhu2024vision+}. For example, in emotion recognition tasks, Li et al. \cite{10076804,9863920,10577436,11049910} have effectively leveraged textual and acoustic to substantially enhance the learned representations. In action recognition, Gao et al. \cite{gao2020listen} have leveraged audio as a preview to eliminate redundant visual information. In recent years, an increasing number of studies have focused on the challenges of multimodal learning in real-world scenarios \cite{zhang2024multimodal,zhang2023provable,9767641,10269037,10224356,10261246}, among which modality imbalance \cite{wang2020makes,wu2022characterizing} has emerged as a critical issue. This paper proposes Adaptive Intra-Network Modulation (AIM), which effectively addresses the modality imbalance problem by modulating the training pace of each modality and stimulating the slow-performing components within the network.

\subsection{Modality Imbalance}
Recent studies have shown that due to discrepancies in performance and convergence speed, jointly optimizing different modalities with a uniform objective often leads to the modality imbalance problem, where certain modalities are suppressed by others \cite{wang2020makes,sun2021learning}. To address this issue, many methods modulate the training pace of different modalities to achieve balanced multimodal learning \cite{peng2022balanced,fan2023pmr,wei2024fly,wei2024diagnosing}. For instance, Peng et al. \cite{peng2022balanced} have introduced on-the-fly gradient modulation to adaptively control the optimization of each modality. Wei et al. \cite{wei2024diagnosing} have proposed a diagnosing and re-learning method that re-initialize each unimodal encoder based on their learning state. Despite their effectiveness, these methods perform modulation at the granularity of entire unimodal networks, ignoring the differences in optimization pace across parameters and layers within the networks, and therefore limiting modulation effectiveness. To address this, we propose a modulation strategy that adaptively stimulates parameters and layers with slower optimization pace within each modality network, thereby improving balanced multimodal learning.

\begin{figure*}[t]
\centering
\includegraphics[width=0.9\textwidth]{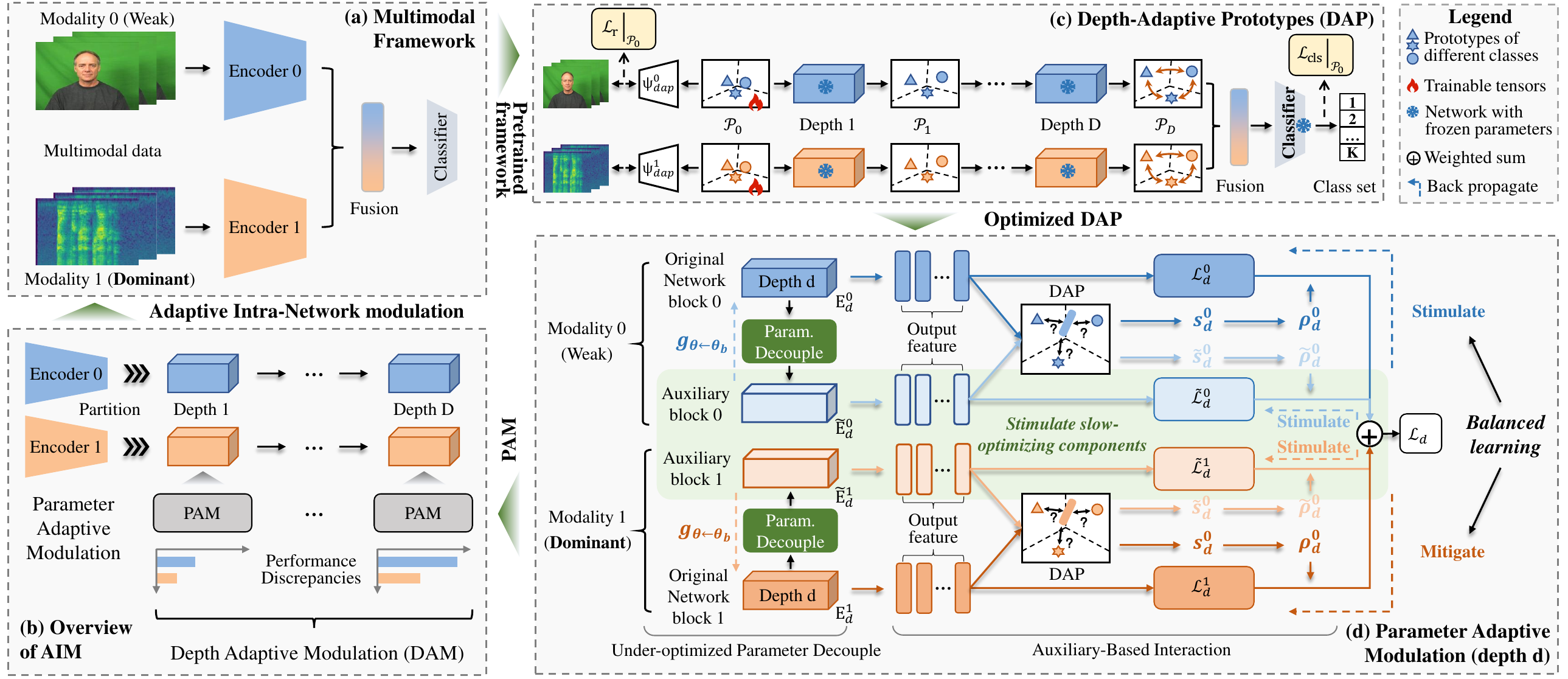} 
\caption{An overview of the proposed Adaptive Intra-Network Modulation (AIM). (a) The multimodal framework; (b) Overview of the of AIM including Parameter-Adaptive Modulation (PAM) and Depth-Adaptive Modulation (DAM); (c) The Depth-Adaptive Prototypes (DAP); (d) Detailed implementation of PAM at depth $d$. The implementation of the parameter decoupling mechanism (green block) is presented in Fig. \ref{fig_param_decouple}. Without loss of generality, this figure illustrates the case of two modalities, with blue and orange representing different modalities, and Modality 1 set as the dominant modality.}
\label{fig_framework}
\end{figure*}

\section{Proposed Approach}
\subsection{Multimodal Framework} \label{sec:mf}
The proposed Adaptive Intra-Network Modulation (AIM) focuses on the joint-training multimodal framework illustrated in Fig. \ref{fig_framework}(a), following previous works \cite{peng2022balanced,fan2023pmr,wei2024diagnosing,wei2024fly}. Consider a multimodal dataset consisting of \(N\) samples $\{x_1,...,x_N\}$ from \(K\) classes, each sample $x_i=\{x_i^1,...,x_i^M\}$ with label $y_i$ comprises data from \(M\) modalities. The framework processes each sample $x_i$ through modality-specific encoders $\{\textbf{E}^m\}_{m=1}^M$ to extract unimodal feature representations. These features are then integrated via a fusion strategy $\Theta_{\mathcal{F}}$, yielding a combined multimodal feature that is subsequently passed to a classifier $\Theta_{\mathcal{C}}$ to output the prediction $\hat{y}_i$. The calculation of the multimodal framework $\Theta_{\mathcal{M}}(\cdot)$ can be formulated as follows:
\begin{align}
    \hat{y}_i=\Theta_{\mathcal{M}}(x_i)=\Theta_{\mathcal{C}}(\Theta_{\mathcal{F}}(\textbf{E}^1(x_i^1),...,\textbf{E}^M(x_i^M))).
    \label{eq:theta-M}
\end{align}
The multimodal framework is trained by minimizing $\mathcal{L}$:
\begin{align}
    \mathcal{L}=CE(y,\hat{y}).
    \label{eq:l}
\end{align}
$CE(\cdot,\cdot)$ is the cross-entropy loss, $y=\{y_i\}_{i=1}^N, \hat{y}=\{\hat{y}_i\}_{i=1}^N$.

Our proposed AIM is agnostic to the choice of encoder backbones and fusion strategies. Experimental results presented in Sec. \ref{sec:val-gen} demonstrate its effectiveness across different fusion methods and backbone architectures.

\subsection{Overview of Adaptive Intra-Network Modulation (AIM)}

We propose Adaptive Intra-Network Modulation (AIM) to address the issue of modality imbalance in the joint training framework while fully exploit the learning potential of each modality network. To achieve this, we first propose depth-adaptive prototypes (DAP) to enable effective performance estimation across different components within the network. Subsequently, a training framework comprising parameter-adaptive modulation (PAM) and depth-adaptive modulation (DAM) is introduced. Specifically, the encoder of each modality \(m\) is first divided into \(D\) sequential blocks arranged from shallow to deep, where the block at depth \(d\) is denoted as \(\textbf{E}_d^m\). In this work, ResNet18 is used as the backbone for each modality encoder, with each of its four residual stages treated as a separate block. At each depth, PAM modulates the learning of all modality network blocks. It leverages DAP to assess the optimization status and adaptively stimulates the learning of under-optimized parameters in network block of each modality, thereby achieving balanced and thorough learning across modalities. DAM further adaptively aggregate the modulation losses of PAM based on modality performance discrepancy at each depth to strengthen PAM on layers with higher imbalance, thereby enhance balanced multimodal learning. DAP, PAM, and DAM are described in detail in Sec. \ref{sec:pap}, \ref{sec:pdm}, and \ref{sec:ddm}, respectively.

\subsection{Depth-Adaptive Prototypes (DAP)} \label{sec:pap}
To enable intra-network modulation, AIM requires evaluation of loss and performance of different layers in each modality network. Previous studies \cite{peng2022balanced,fan2023pmr,wei2024fly,wei2024diagnosing} typically use ground-truth labels to evaluate the overall performance of a network, regarding the labels as the optimal network prediction. However, different components within the network exhibit varying optimization capacities, i.e., their optimal predictions may deviate from the ground truth to different extents. Motivated by this consideration, we propose to use depth-adaptive prototypes (DAP) instead of ground-truth labels for evaluating different components within the network. 

Specifically, for the multimodal framework described in Sec. \ref{sec:mf}, DAP assigns \(K\) feature vectors (prototypes) at the output of each depth \(d\) in the network of each modality \(m\), denoted as \(\mathcal{P}_d^m=\{(p_d^m)_k\}_{k=1}^K\), where \(K\) is the number of classes. Each prototype \((p_d^m)_k\) has the same dimensionality as the corresponding depth output and represents the optimal latent feature that would ultimately be classified into class \(k\) through the subsequent network layers. As shown in Fig.\ref{fig_framework}(c), we first pre-train the multimodal framework and freeze all its parameters. Then, we initialize \(K\) learnable feature vectors at the input of each modality encoder $\textbf{E}^m$, denoted as $\mathcal{\Tilde{P}}_0^m=\{(\Tilde{p}_0^m)_k\}_{k=1}^K$. Two losses, the classification loss \(\mathcal{L}_{\text{cls}}\) and the reconstruction loss \(\mathcal{L}_{\text{r}}\), are then defined to train \(\{\mathcal{\Tilde{P}}_0^m\}_{m=1}^M\). \(\mathcal{L}_{\text{cls}}\) enhances the classification accuracy of the prototypes by minimizing the discrepancy between the outputs of \(\{\mathcal{\Tilde{P}}_0^m\}_{m=1}^M\) after passing through the entire multimodal framework and the class set $\mathcal{Y}=[0,\cdots,K{-}1]$, which is defined as follows:
\begin{align}
    \mathcal{L}_{\text{cls}}\mid_{\mathcal{\Tilde{P}}_0}=CE(\Theta_{\mathcal{M}}(\mathcal{\Tilde{P}}_0), \mathcal{Y}).
    \label{eq:l-proto}
\end{align}
$CE(\cdot,\cdot)$ is the cross-entropy loss. \(\mathcal{L}_{\text{r}}\) ensures the information consistency of the prototypes with the input data by minimizing the discrepancy between the decoded $\mathcal{\Tilde{P}}_0^m$ and the mean feature vectors of each class in the input data of each modality $m$, which is defined as follows:
\begin{align}
    \mathcal{L}_{\text{r}}\mid_{\mathcal{\Tilde{P}}_0}=\frac{1}{M\times K}\sum_{m=1}^M{\sum_{k=1}^K{\|\Psi_{dap}^m((p_0^m)_k)-\bar{X}^{m,k}\|_{\mathrm{F}}^2}}.
\end{align}
$\Psi_{dap}^m(\cdot)$ is the decoder to decode $(p_d^m)_k$ for modality $m$. $\bar{X}^{m,k}$ denotes the mean feature vector of class $k$ in modality $m$ of the dataset. $\|\cdot\|_{\mathrm{F}}^2$ is the Frobenius norm. Therefore, by minimizing $\mathcal{L}_{\text{cls}}$ and $\mathcal{L}_{\text{r}}$, $\mathcal{\Tilde{P}}_0$ is optimized toward $\mathcal{P}_0$, which can be expressed as follows:
\begin{align}
\mathcal{P}_0=\mathop{\arg\min}\limits_{\mathcal{\Tilde{P}}_0}\mathcal{L}_{\text{cls}}|_{\mathcal{\Tilde{P}}_0}+\mathcal{L}_{\text{r}}|_{\mathcal{\Tilde{P}}_0}.
    \label{eq:pap_opt}
\end{align}
Then, the prototypes at each depth $d$ can be obtained through forward propagation, as expressed by the following equation:
\begin{align}
    \mathcal{P}_d^m=\textbf{E}_{\sim d}^m(\mathcal{P}_0^m),\quad \forall m\in[1,M],\; \forall d\in[1,D],
    \label{eq:forward-pd}
\end{align}
$\mathcal{P}_d^m$ denotes the prototypes located at the output of depth \(d\) in the encoder \(\textbf{E}^m\) of modality \(m\). $\textbf{E}_{\sim d}^m$ denotes the sub-network composed of the first $d$ blocks of the encoder $\textbf{E}^m$. In the following intra-network modulation, we use \(\mathcal{P}_d^m\) to estimate the performance and loss for modality \(m\) at depth \(d\).

\begin{figure}[t]
\centering
\includegraphics[width=0.9\columnwidth]{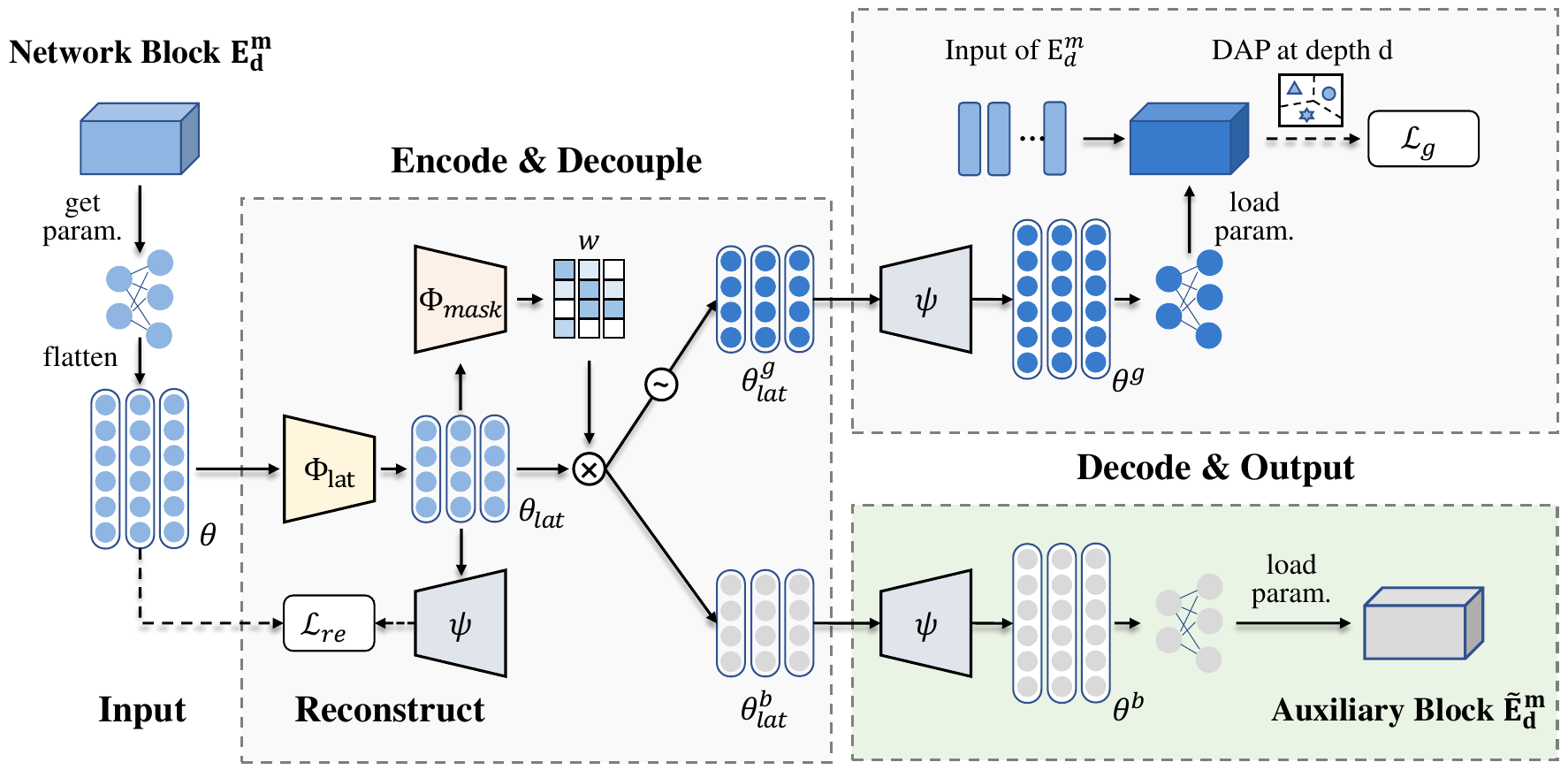} 
\caption{The detailed implementation of the parameter decoupling mechanism applied on the network block $\textbf{E}_d^m$.}
\label{fig_param_decouple}
\end{figure}

\subsection{Parameter-Adaptive Modulation (PAM)} \label{sec:pdm}
At each depth of the multimodal framework, the Parameter-Adaptive Modulation (PAM) is proposed to adaptively balance the learning and stimulate under-optimized parameters across modalities based on their performance. As shown in Fig. \ref{fig_framework}(c), a \textbf{\textit{parameter decoupling mechanism}} is first introduced for under-optimized parameter decoupling. A \textbf{\textit{auxiliary-based interaction}} is then applied for adaptive balancing and stimulation.


\subsubsection{Parameter Decoupling Mechanism}
The implementation of the parameter decoupling mechanism is illustrated in Fig. \ref{fig_param_decouple}. For the network block at depth $d$ of modality $m$ encoder, denoted as $\textbf{E}_d^m$, its parameters $\theta$ are first flattened, which are then mapped into the latent space by encoder $\Phi_{lat}$ to obtain $\theta_{lat}$: $\theta_{lat}=\Phi_{lat}(\theta)$. Based on $\theta_{lat}$, $\Phi_{mask}$ learns a mask $w$ of the same dimension: $w=\sigma(\Phi_{mask}(\theta_{lat}))$, where $\sigma$ is the sigmoid function. The mask $w$ is employed to select a subset from $\theta_{lat}$ that is under-optimized with poor performance: $\theta_{lat}^b=\theta_{lat}\odot w$. $\odot$ is the element-wise multiplication. The selected $\theta_{lat}^b$ is then decoded by $\Psi$, yielding $\theta^b$: $\theta^b=\Psi(\theta_{lat}^b)$, which is finally loaded $\theta^b$ into the network structure identical to $\textbf{E}_d^m$ and obtains a new block $\Tilde{\textbf{E}}_d^m$. In summary, the decoupling of $\theta_b$ from the input $\theta$ using the parameter decoupling mechanism $\mathcal{F}_{pdm}$ is given in Eq. (\ref{eq:fpdm}).
\begin{align}
    \theta_b=\mathcal{F}_{pdm}(\theta)=\Psi(\Phi_{lat}(\theta)\odot\sigma(\Phi_{mask}(\Phi_{lat}(\theta)))).
    \label{eq:fpdm}
\end{align}
To effectively train the parameter decoupling mechanism, we design a complementary branch opposite to the computation of $\theta^b$. Specifically, we use $(1-w)$ to select $\theta_{lat}^g$ from $\theta_{lat}$, formulated as $\theta_{lat}^g=\theta_{lat}\odot(1-w)$. Then, the decoder $\Psi$ maps $\theta_{lat}^g$ to $\theta^g$, which is also loaded into the network structure identical to $\textbf{E}_d^m$ to form another network block $\textbf{E}|_{\theta^g}$. We minimize the loss $\mathcal{L}_g$ of $\textbf{E}|_{\theta^g}$ against corresponding DAP $\mathcal{P}_d^m$ as well as the reconstruction loss $\mathcal{L}_{re}$, which is formulated in Eq. (\ref{eq:L_pdm}).
\begin{align}
    &\mathcal{L}_{pdm}=\mathcal{L}_{g}+\mathcal{L}_{re} \notag \\
    &=\mathbb{E}_{\textbf{p}(I^m_d,y)}\left[-log\frac{exp(-\delta\left[\textbf{E}|_{\theta_g}(I_d^m),(p_d^m)_y\right])}{\sum_{k=1}^K{exp(-\delta\left[\textbf{E}|_{\theta_g}(I_d^m),(p_d^m)_k\right])}}\right] \notag \\
    &+\bigl\|\theta-\Psi(\theta_{lat})\bigr\|_{\mathrm{F}}^2.
    \label{eq:L_pdm}
\end{align}
$I_d^m$ represents the input feature vectors of network block $\textbf{E}_d^m$, i.e., the output of $\textbf{E}_{d-1}^m$ and equals to $x^m$ when $d=0$. $y$ is their corresponding ground-truth label. $\textbf{p}(\cdot,\cdot)$ is the joint distribution of two variables. $\delta[\cdot,\cdot]$ denotes the Euclidean distance between two vectors. $(p_d^m)_{y_i}$ is the DAP for class $y_i$ of modality $m$ at depth $d$. Minimizing $\mathcal{L}_g$ ensures the selected $\theta^g$ are well-performing parameters, while its complementary $\theta^b$ exhibits degraded performance, while $\mathcal{L}_{re}$ further constrains the learning of all the encoders and the decoder $\Psi$.

\subsubsection{Auxiliary-Based Interaction} \label{sec:wpwis}
For each original network block \(\textbf{E}_d^m\) at depth \(d\), a corresponding block $\Tilde{\textbf{E}}_d^m$ with under-optimized parameters, termed the Auxiliary Block, is derived through the Parameter Decoupling Mechanism. Subsequently, all original network blocks and Auxiliary Blocks are jointly trained, with their participation adaptively adjusted based on performance. Specifically, the performance and loss of $\textbf{E}_d^m$ and $\Tilde{\textbf{E}}_d^m$ are first evaluated via DAP $\mathcal{P}_d^m$:
\begin{align}
    s^m_d&=\mathbb{E}_{\textbf{p}(I^m_d,y)}\left[\frac{exp\left(-\delta\left[\textbf{E}_d^m(I^m_d),(p_d^m)_y\right]\right)}{\sum_{y^{\prime}}{exp\left(-\delta\left[\textbf{E}_d^m(I^m_d),(p_d^m)_{y^{\prime}}\right]\right)}}\right],\label{eq:smd}\\
    \mathcal{L}_d^m&=\mathbb{E}_{\textbf{p}(I^m_d,y)}\left[-log\frac{exp(-\delta\left[\textbf{E}_d^m(I_d^m),(p_d^m)_y\right])}{\sum_{k=1}^K{exp(-\delta\left[\textbf{E}_d^m(I_d^m),(p_d^m)_k\right])}}\right]. \label{eq:l-d-m}
\end{align}
$s^m_d$ and $\mathcal{L}_d^m$ are the performance and loss of $\textbf{E}_d^m$. The performance $\Tilde{s}_d^m$ and loss $\mathcal{\Tilde{L}}^{m}_d$ of $\Tilde{\textbf{E}}_d^m$ can be calcuated by replacing $\textbf{E}_d^m$ with $\Tilde{\textbf{E}}_d^m$ in Eq. (\ref{eq:smd}) and (\ref{eq:l-d-m}). Subsequently, $\{\mathcal{L}_d^m\}_{m=1}^M$ and $\{\mathcal{\Tilde{L}}_d^m\}_{m=1}^M$ are integrated based on $\{s_d^m\}_{m=1}^M$ and $\{\Tilde{s}_d^m\}_{m=1}^M$, which can be formulated as follows:
\begin{align}
    \mathcal{L}_d=\sum_{m=1}^M \left[\rho_d^m\mathcal{L}_d^m+\Tilde{\rho}_d^m\mathcal{\Tilde{L}}_d^m\right], \label{eq:l-d}
\end{align}
where
\begin{align}
    \rho_d^m=\frac{\exp(-s_d^m)}{\sum_{m=1}^M{\exp(-s_d^m)+\exp(-\Tilde{s}_d^m)}}, \label{eq:rho-dm}
\end{align}
and $\Tilde{\rho}_d^m$ can be computed by replacing $s_d^m$ in the numerator of Eq. (\ref{eq:rho-dm}) with $\Tilde{s}_d^m$. By using $\mathcal{L}_d$ for joint training, multimodal learning is effectively balanced by reducing the participation of high-performing original blocks in dominant modalities while enhancing that of weak modalities. More importantly, the involvement of the Auxiliary Blocks stimulates the learning of under-optimized parameters in each modality during joint training. In particular, parameter updates within the Auxiliary Blocks can be backpropagated to their corresponding network original blocks, as formulated in Eq. (\ref{eq:opt-full}).
\begin{align}
    g_{\theta\leftarrow\theta_b}=J_{\mathcal{F}_{pdm}}(\theta)^{\!T}\cdot g_{\theta_b}=J_{\mathcal{F}_{pdm}}(\theta)^{\!T}\cdot \Tilde{\rho}_d^m\frac{\partial \mathcal{\Tilde{L}}_d^m}{\theta_b},
    \label{eq:opt-full}
\end{align}
$\theta$ and $\theta_b$ represents the network parameters of $\textbf{E}_d^m$ and its corresponding auxiliary block $\Tilde{\textbf{E}}_d^m$, respectively. $g_{\theta\leftarrow\theta_b}$ denotes the gradient back-propagated to $\theta$ from the update of $\theta_b$. $g_{\theta_b}$ is the gradient of $\theta_b$. $J_{\mathcal{F}_{pdm}}(\theta)$ is the the Jacobian matrix of the parameter decoupling mechanism $\mathcal{F}_{pdm}$ at $\theta$. $\Tilde{\rho}_d^m$ is the fixed weight in Eq. (\ref{eq:l-d}). Therefore, \textbf{for dominant modalities}, even though direct training of their original network blocks is reduced, their optimization can still be enhanced via Eq. (\ref{eq:opt-full}), preventing excessive mitigation of their learning. \textbf{For weak modalities}, the suppression they experience is eliminated, and training focuses on the parameters with slower optimization pace to avoid insufficient stimulation.

\begin{algorithm}[t]
\caption{Multimodal training with AIM.}
\label{alg:algorithm}
\textbf{Input}: the starting epoch of modulation $E$, the total training epochs $E_T$, and the randomly initialized learnable tensor $\Tilde{\mathcal{P}}_0$.

\begin{algorithmic}[1] 
\FOR{$e=0,...,E-1$}
\STATE Train the multimodal framework and the parameter decoupling mechanism jointly using Eqs. (\ref{eq:l}) and (\ref{eq:L_pdm}). Freeze the multimodal framework and optimize DAP using Eq. (\ref{eq:pap_opt}).
\ENDFOR
\FOR{$e=E,...,E_{T-1}$}
\STATE Decouple the parameters of each network block $\textbf{E}_d^m$ using Eq. (\ref{eq:fpdm}) and obtain the corresponding auxiliary block $\Tilde{\textbf{E}}_d^m$;
\STATE Calculate $s_d^m$ using Eq. (\ref{eq:smd});
\STATE Calculate $\alpha_d$ using Eq. (\ref{eq:alpha-d});
\STATE Calculate $\mathcal{L}_d^m, \Tilde{\mathcal{L}}_d^m$ and $\mathcal{L}_d$ using Eqs. (\ref{eq:l-d-m}) and (\ref{eq:l-d});
\STATE Calculate $\mathcal{L}_{mod}$ using Eq. (\ref{eq:l_mod});
\STATE Jointly update multimodal framework and DAP using $\mathcal{L}_{mod}$ and Eq. (\ref{eq:pap_opt}).
\ENDFOR
\end{algorithmic}
\label{alg:aim}
\end{algorithm}

\subsection{Depth-Adaptive Modulation} \label{sec:ddm}
The parameter-adaptive modulation (PAM) balances and stimulates multimodal learning at each depth $d$ via $\mathcal{L}_d$. However, differences in optimization paces across depths of modality networks also lead to varying levels of imbalance at different depths within the multimodal framework during training. Therefore, depth-adaptive modulation further adjusts the strength of PAM according to the imbalance level at each depth. Specifically, we first compute the performance discrepancy between modalities to evaluate the level of modality imbalance at each depth $d$. Considering that performance values may vary substantially across depths, we employ the Coefficient of Variation \cite{abdi2010coefficient} to calculate the discrepancy, which is formulated in Eq. (\ref{eq:alpha-d}).
\begin{align}
    \alpha_d=\frac{\sqrt{\frac{1}{M}\sum_{m=1}^M(s_d^m-\Bar{s}_d)^2}}{\Bar{s}_d},
    \label{eq:alpha-d}
\end{align}
where $s_d^m$ is calculated from Eq. (\ref{eq:smd}) using DAP. $\Bar{s}_d=\frac{1}{M}\sum_{m=1}^M{s_d^m}$. $\alpha_d$ is the performance discrepancy between modalities that measures the imbalance level in depth $d$. Subsequently, the multimodal modulation losses $\{\mathcal{L}_d\}_{d=1}^D$ at different depths are weighted by $\{\alpha_d\}_{d=1}^D$ and integrated to form the total modulation loss of AIM, as formulated in Eq.~(\ref{eq:l_mod}).
\begin{align}
    \mathcal{L}_{mod} &= \sum_{d=1}^D{\alpha_d\cdot \mathcal{L}_d}.
    \label{eq:l_mod}
\end{align}
In this way, DAM enhances balanced multimodal learning by emphasising modulation on depths with greater imbalance.

\subsection{Training and Inference} \label{sec:t&i}
The training of the multimodal framework with AIM is demonstrated in Algorithm \ref{alg:aim}. The value of the starting epoch of modulation $E$ will be analyzed in Sec. \ref{sec:e-analysis}. During testing, the well-trained multimodal framework is directly used for prediction.

\section{Experiments}

\subsection{Experimental Setups} \label{sec:exp-set}
\subsubsection{Datasets} \label{sec:dataset} Experiments are conducted on four commonly used multimodal datasets in previous methods \cite{peng2022balanced,fan2023pmr,wei2024fly,wei2024diagnosing}. \textbf{(1) CREMA-D} \cite{cao2014crema} is an emotion recognition dataset comprising audio and visual modalities covering six emotion categories: angry, happy, sad, neutral, disgust, and fear. \textbf{(2) Kinetics-Sounds (KS)} \cite{arandjelovic2017look} is an action recognition dataset including 31 selected categories from the Kinetics dataset \cite{kay2017kinetics} with two modalities, audio and visual. \textbf{(3) UCF-101} \cite{soomro2012ucf101} is an action recognition dataset comprising 101 human action categories and two modalities: RGB and optical flow. \textbf{(4) CMU-MOSI} \cite{zadeh2016mosi} is a sentiment analysis dataset comprising 3,228 YouTube videos, each containing audio, visual, and textual modalities. \textbf{Please refer to the Supplemental Materials for full description of datasets.}

\subsubsection{Compared methods} \label{sec:comp-method}
To evaluate the effectiveness of the proposed Adaptive Intra-Network Modulation (AIM) in addressing the problem of modality imbalance, several recent studies are introduced for comparison, including Gradient-Blending (\textbf{G-Blending}) \cite{wang2020makes}, On-the-fly Gradient Modulation with Generalization Enhancement (\textbf{OGM-GE}) \cite{peng2022balanced}, \textbf{Greedy} \cite{wu2022characterizing}, Prototypical Modality Rebalance (\textbf{PMR}) \cite{fan2023pmr}, Adaptive Gradient Modulation (\textbf{AGM}) \cite{li2023boosting}, Multimodal Learning with Alternating Unimodal Adaptation (\textbf{MLA}) \cite{zhang2024multimodal}, Diagnosing and Re-learning (\textbf{D\&R}) \cite{wei2024diagnosing}, and On-the-fly Prediction Modulation and On-the-fly Gradient Modulation (\textbf{OPM\&OGM}) \cite{wei2024fly}. Joint training using concatenation fusion is adopted as the baseline for studying the modality imbalance problem, following previous works \cite{peng2022balanced,fan2023pmr,wei2024diagnosing,wei2024fly}.

\subsubsection{Evaluation metrics} \label{sec:eval-metrics} In the experiments, accuracy (ACC) and macro-averaged F1 score (MacroF1) are employed to evaluate the performance of different methods following previous works \cite{peng2022balanced,fan2023pmr,wei2024fly,wei2024diagnosing}.

\subsubsection{Implementation Details}
The proposed method is built upon the multimodal framework described in Sec. \ref{sec:mf}. Following previous works \cite{peng2022balanced,wei2024diagnosing}, ResNet18 is adopted as the backbone for CREMA-D and Kinetic Sounds, and the models are trained from scratch. For UCF-101, ResNet18 is also used as the backbone and is pretrained on ImageNet. For CMU-MOSI, a transformer-based network serves as the backbone, with models trained from scratch. In the parameter decoupling module, both the encoder and decoder consist of a single fully connected layer. The encoder takes as input the number of parameters at each network layer and maps them to a 512-dimensional latent space. The decoder takes a 512-dimensional vector as input and outputs the same number of parameters as in the corresponding network layer. Dataset preprocessing follows previous works \cite{peng2022balanced,wei2024diagnosing}. All models are trained using SGD with a momentum of 0.9 and a learning rate of 1e-3, and all experiments are conducted on one NVIDIA RTX A6000.


\begin{table}[t]
\caption{Comparison with methods for balanced multimodal learning on the CREMA-D, Kinetics Sounds, and UCF-101 datasets. Bold and underline represent the best and second best, respectively. Joint-training with concatenation fusion is used as a baseline following previous works.}
\begin{center}
\resizebox{\columnwidth}{!}{%
\begin{tabular}{c|cc|cc|cc}
\hline
\multirow{3}{*}{Method} & \multicolumn{2}{c|}{CREMA-D}        & \multicolumn{2}{c|}{Kinetics-Sounds} & \multicolumn{2}{c}{UCF-101}  \\ 
                        & \multicolumn{2}{c|}{(Audio+Visual)}   & \multicolumn{2}{c|}{(Audio+Visual)}    & \multicolumn{2}{c}{(RGB+Optical Flow)} \\ 
                        & \multicolumn{1}{c}{ACC} & Macro-F1 & \multicolumn{1}{c}{ACC}  & Macro-F1 & \multicolumn{1}{c}{ACC}  & Macro-F1  \\ \hline
Joint-training          & \multicolumn{1}{c}{67.47}   & 67.80        & \multicolumn{1}{c}{65.04}    & 65.12        & \multicolumn{1}{c}{67.34}    & 66.93 \\ \hline
G-Blending \cite{wang2020makes}                      & \multicolumn{1}{c}{69.89}   & 70.41        & \multicolumn{1}{c}{68.60}    & 68.64        & \multicolumn{1}{c}{72.15}    & 71.31\\ 
OGM-GE \cite{peng2022balanced}                      & \multicolumn{1}{c}{68.95}   & 69.39        & \multicolumn{1}{c}{67.15}    & 66.93        & \multicolumn{1}{c}{71.84}    & 70.74 \\ 
Greedy \cite{wu2022characterizing}                      & \multicolumn{1}{c}{68.37}   & 68.46        & \multicolumn{1}{c}{65.72}    & 65.80        & \multicolumn{1}{c}{70.61}    & 70.46 \\ 
PMR \cite{fan2023pmr}                      & \multicolumn{1}{c}{68.55}   & 68.99        & \multicolumn{1}{c}{65.62}    & 65.36        & \multicolumn{1}{c}{72.25}    & 71.16 \\ 
AGM \cite{li2023boosting}                      & \multicolumn{1}{c}{70.16}   & 70.67        & \multicolumn{1}{c}{66.50}    & 66.49        & \multicolumn{1}{c}{72.31}    & 71.52 \\ 
MLA \cite{zhang2024multimodal}                      & \multicolumn{1}{c}{\underline{79.70}}   & \underline{79.94}        & \multicolumn{1}{c}{\underline{71.32}}    & \underline{71.23}        & \multicolumn{1}{c}{\underline{74.51}}    & \underline{73.44}  \\ 
D\&R \cite{wei2024diagnosing}                      & \multicolumn{1}{c}{75.13}   & 76.00        & \multicolumn{1}{c}{69.10}    & 69.39        & \multicolumn{1}{c}{73.12}    & 71.45  \\ 
OPM\&OGM \cite{wei2024fly}                      & \multicolumn{1}{c}{75.10}   & 75.91        & \multicolumn{1}{c}{68.00}    & 68.12        & \multicolumn{1}{c}{72.36}    & 71.62     \\ \hline
Ours                       & \multicolumn{1}{c}{\textbf{81.32}}   & \textbf{82.00}        & \multicolumn{1}{c}{\textbf{72.40}}    & \textbf{72.27}        & \multicolumn{1}{c}{\textbf{77.43}}    & \textbf{78.12}  \\ \hline
\end{tabular}%
}
\end{center}
\label{tab:sota}
\end{table}

\subsection{Comparison With Baselines}\label{sec:comp-sota}
\subsubsection{Comparison with the Joint-Training Baseline} As shown in Table \ref{tab:sota}, all methods for balanced multimodal learning, including ours, outperform the joint-training baseline across all datasets. Our method achieves particularly notable gains. On CREMA-D, it improves ACC and Macro-F1 by 13.85\% and 14.20\%, respectively, while on Kinetics-Sounds, the improvements are 7.36\% and 7.15\%. On UCF-101,we achieves 10.09\% higher ACC and 11.19\% higher Macro-F1 compared with joint training. These results confirm the presence of the modality imbalance and underscore the importance of modulating each unimodal learning to achieve balanced learning.

\subsubsection{Comparison with State-of-the-Art Methods for Balanced Multimodal Learning} Compared with recent studies for balanced multimodal learning, our method consistently achieves superior performance across all datasets. On the CREMA-D dataset, it outperforms the state-of-the-art method MLA by 1.62\% in ACC and 2.06\% in Macro-F1, and exceeds other advanced approaches such as D\&R by at least 6.13\% and 6\%, respectively. On the Kinetics-Sounds dataset, our method yields at least a 3.3\% gain in ACC over leading methods. On the UCF-101 dataset, our method shows improvements of 2.92\% in ACC and 4.68\% in Macro-F1 relative to MLA. This performance gain may stem from our modulation strategy, which better exploits each modality's learning potential.

\subsection{Ablation study}\label{sec:abl-study}
Ablation studies are conducted to evaluate the contributions of parameter-adaptive modulation (PAM) and depth-adaptive modulation (DAM) of AIM in enhancing multimodal performance. Specifically, we introduce two degraded variants, ``w/o PAM'' and ``w/o DAM''. The ``w/o PAM'' variant removes the parameter decoupling module and auxiliary block, computing the multimodal modulation loss $\mathcal{L}_d$ as a weighted sum of the losses of original network blocks of all the modalities. The ``w/o DAM'' variant sets all depth-wise weights $\alpha_d$ in Eq. (\ref{eq:l_mod}) to 1. Table \ref{tab:abl} presents a comparison between the two ablated variants, ``w/o PAM'' and ``w/o DAM'', and the full AIM model. The results demonstrate that both ablations improve performance over the joint training baseline across multiple datasets, as both retain a certain degree of modulation. However, their performance is inferior to that of the complete AIM, indicating that both PAM and DAM are crucial for enhancing balanced multimodal learning. Furthermore, compared to ``w/o DAM'', the removal of PAM results in a larger performance decline relative to AIM and achieves a smaller gain over joint training. The main reason is that PAM not only balances learning across modalities but also significantly enhances multimodal performance by fully stimulating the learning of each modality network, while DAM further refines the efficiency of modulation on top of it.


\begin{table}[t]
\caption{Ablation study validating the effectiveness of parameter-adaptive modulation (\textbf{PAM}) and depth-adaptive modulation (\textbf{DAM}) in AIM.}
\begin{center}
\resizebox{\columnwidth}{!}{%
\begin{tabular}{c|cc|cc|cc}
\hline
\multirow{3}{*}{Method} & \multicolumn{2}{c|}{CREMA-D}        & \multicolumn{2}{c|}{Kinetics-Sounds} & \multicolumn{2}{c}{UCF-101}  \\ 
                        & \multicolumn{2}{c|}{(Audio+Visual)}   & \multicolumn{2}{c|}{(Audio+Visual)}    & \multicolumn{2}{c}{(RGB+Optical Flow)} \\ 
                        & \multicolumn{1}{c}{ACC} & Macro-F1 & \multicolumn{1}{c}{ACC}  & Macro-F1 & \multicolumn{1}{c}{ACC}  & Macro-F1  \\ \hline
Joint-training          & \multicolumn{1}{c}{67.47}   & 67.80        & \multicolumn{1}{c}{65.04}    & 65.12        & \multicolumn{1}{c}{67.34}    & 66.93   \\ \hline
w/o PAM          & \multicolumn{1}{c}{75.12}   & 75.41        & \multicolumn{1}{c}{68.83}    & 68.36        & \multicolumn{1}{c}{71.43}    & 71.39  \\
w/o DAM                      & \multicolumn{1}{c}{80.14}   & 80.52        & \multicolumn{1}{c}{71.82}    & 71.51        & \multicolumn{1}{c}{75.92}    & 75.52 \\ \hline
AIM                      & \multicolumn{1}{c}{\textbf{81.32}}   & \textbf{82.00}        & \multicolumn{1}{c}{\textbf{72.40}}    & \textbf{72.27}        & \multicolumn{1}{c}{\textbf{77.43}}    & \textbf{78.12}   \\ \hline
\end{tabular}%
}
\end{center}
\label{tab:abl}
\end{table}

\subsection{Validation of Generalizability} \label{sec:val-gen}

In multimodal learning, beyond convolution-based deep networks such as ResNet, more sophisticated transformer-based backbones are also widely adopted. Accordingly, we further conduct experiments with transformer-based backbones to compare various balanced multimodal learning methods. Following previous work \cite{wei2024diagnosing}, we employ the transformer encoder as the backbone of each unimodal encoder on the CMU-MOSI dataset, with model trained from scratch. Specifically, we use a 3-layer transformer encoder \cite{liang2021multibench}, where each layer (comprising multi-head self-attention and feed-forward parameters) is treated as a network block in our method. The results in Table \ref{tab:trans_back} demonstrate that balanced multimodal learning methods are generally less effective on transformer-based backbones, which is consistent with previous findings \cite{wei2024diagnosing}. Nevertheless, our method achieves superior performance compared to others, highlighting its stronger versatility.

More experiments are conducted to validate the effectiveness of AIM across different modality fusion strategies and optimizers. \textbf{Due to space limitations, these results are provided in the Supplemental Materials.}

\begin{table}[t]
\caption{Comparison of different methods using a Transformer-based backbone on the CMU-MOSI dataset. Methods that do not support more than two modalities are excluded from the comparison.}
\begin{center}
\begin{tabular}{c|cc}
\hline
\multirow{3}{*}{Method} & \multicolumn{2}{c}{CMU-MOSI} \\ 
                        & \multicolumn{2}{c}{(Audio+Visual+Text)}  \\ 
                        & \multicolumn{1}{c}{ACC} & Macro-F1  \\ \hline
Joint-training          & \multicolumn{1}{c}{76.96}   & 75.68     \\ \hline
G-Blending \cite{wang2020makes}                      & \multicolumn{1}{c}{77.26}   & 76.27  \\ 
AGM \cite{li2023boosting}                      & \multicolumn{1}{c}{77.26}   & 76.02  \\ 
MLA \cite{zhang2024multimodal}                      & \multicolumn{1}{c}{\underline{78.42}}   & \underline{78.21}   \\ 
D\&R \cite{wei2024diagnosing}                      & \multicolumn{1}{c}{77.99}   & 77.37  \\  \hline
Ours                       & \multicolumn{1}{c}{\textbf{79.51}}   & \textbf{79.37}  \\ \hline
\end{tabular}%
\end{center}
\label{tab:trans_back}
\end{table}

\subsection{Enhanced Unimodal Performance with AIM} \label{sec:val-ww}
As noted in Sec. \ref{sec:intro}, AIM can fully leverage the learning potential of each modality network, effectively enhancing the performance of each unimodal without hindering dominant modalities in the process of stimulating weak ones. To verify this, we observe the performance of each modality on AIM during training and compare it with the corresponding unimodal model, the multimodal joint training baseline, and the advanced method for balanced multimodal learning, D\&R\cite{wei2024diagnosing}. Fig. \ref{fig:winwin} presents the results for the audio and visual modalities on Kinetics-Sounds. The audio modality achieves higher accuracy than the visual modality, indicating that audio is the dominant modality. Moreover, the performance of the dominant audio modality remains similar in both audio-only and joint-training multimodal models, whereas the performance of the weaker visual modality drops substantially under joint training. This suggests that, during joint training, the dominant audio modality suppresses the learning of the weaker visual modality, resulting in an imbalanced learning outcome. D\&R achieves better performance on the weak visual modality compared to joint training, but performs worse on the dominant audio modality. This indicates that, although D\&R effectively alleviates the suppression of the weak modality, it also hinders the learning of the dominant modality to some extent. Also in Table \ref{tab:winwin}, although existing methods generally improve the performance of the weak modality compared to joint training, they compromise the performance of the dominant modality. In contrast, the proposed AIM achieves performance on both the dominant and weak modalities that is comparable to or even outperforms their corresponding unimodal model. This indicates that AIM not only balances learning across modalities but also improves the performance of each unimodal network.

\begin{figure}[t]
\centering
\includegraphics[width=0.85\columnwidth]{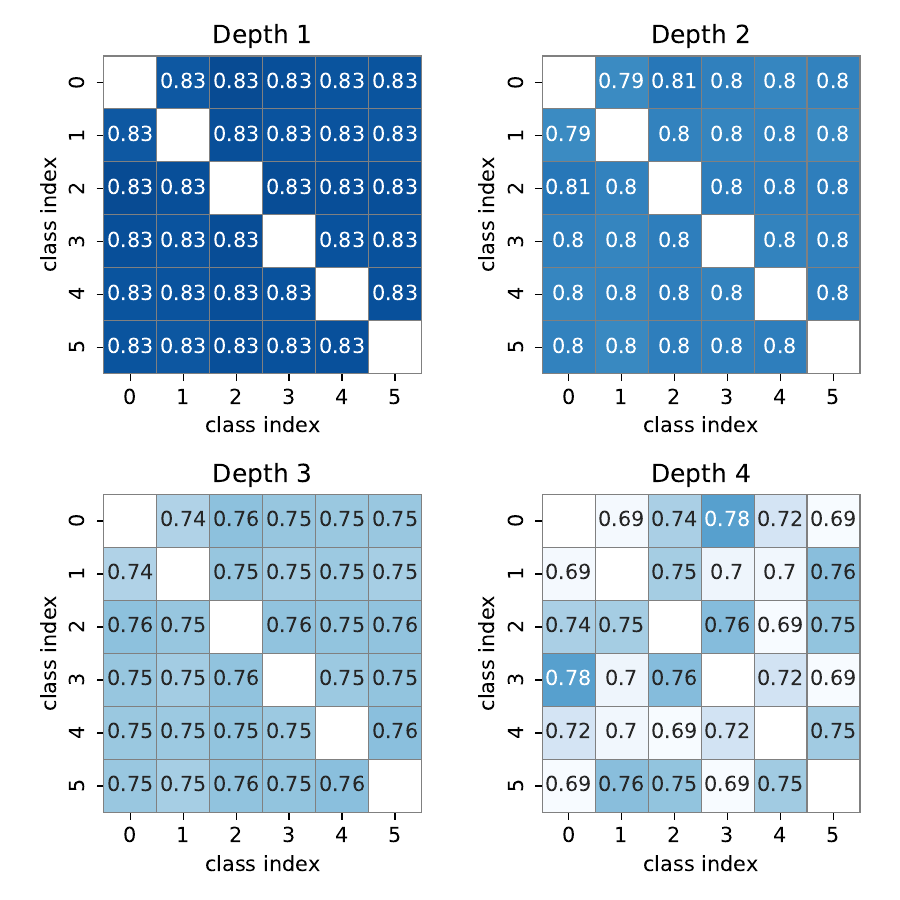} 
\caption{Visualization of inter-class orthogonality of depth-adaptive prototypes at different depths of the audio network on CREMA-D.}
\label{fig:dap-oth}
\end{figure}

\subsection{Discussion on the Depth-Aware Prototypes}
To evaluate the improvement of Depth-Aware Prototypes (DAP) over existing ground-truth label–based performance evaluation methods within AIM, a variant ``AIM-label'' is introduced in the experiments. As implemented in previous works \cite{peng2022balanced,wei2024fly}, ``AIM-label'' employs the classifier to obtain the predicted distribution of each modality at different depths, replacing the computation in Eq. (\ref{eq:smd}) with the probability of the ground-truth class under the predicted distribution. Moreover, it replaces Eq. (\ref{eq:l-d-m}) with the cross-entropy loss between the prediction and the ground-truth labels. The comparison between AIM and ``AIM-label'' in Table \ref{tab:pap-vs-label} shows that both ``AIM-label'' and AIM improve multimodal performance compared to joint training, while AIM with DAP achieves a greater improvement than ``AIM-label''. This indicates that the use of DAP supports more effective modulation in AIM.

\begin{table}[t]
\caption{Performance of ``AIM-label'' and AIM on different datasets.}
\begin{center}
\begin{tabular}{c|c|c|c|c}
\hline
\multirow{2}{*}{Method} & \multicolumn{2}{c|}{CREMA-D}        & \multicolumn{2}{c}{Kinetics-Sounds}  \\ 
                        & \multicolumn{1}{c}{ACC} & Macro-F1 & \multicolumn{1}{c}{ACC}  & Macro-F1 \\ \hline
Joint-training                  & \multicolumn{1}{c}{67.47} & 67.80 & \multicolumn{1}{c}{65.04} & 65.12 \\ \hline
AIM-label & \multicolumn{1}{c}{80.11} & 80.24 & \multicolumn{1}{c}{71.05} & 70.82 \\ 
AIM & \multicolumn{1}{c}{\textbf{81.32}} & \textbf{82.00} & \multicolumn{1}{c}{\textbf{72.40}} & \textbf{72.27} \\ \hline
\end{tabular}
\end{center}
\label{tab:pap-vs-label}
\end{table}

To further evaluate the adaptivity of DAP to the optimization capacities at different depths, we visualize the inter-class orthogonality of the prototypes at each depth. Fig. \ref{fig:dap-oth} shows the heatmap of matrices $\{\mathcal{O}_d^m\}_{d=1}^D$ calculated via Eq. (\ref{eq:orth}) on the CREMA-D dataset.
\begin{align}
    \mathcal{O}_d^m\in\mathbb{R}^{K\times K}=\langle S(\mathcal{P}_d^m), S(\mathcal{P}_d^m)^{\!T}\rangle.
    \label{eq:orth}
\end{align}
\(\langle\cdot,\cdot\rangle\) denotes the cosine similarity between two tensors. \(S(\cdot)\) stacks a set of tensors into a single tensor. The superscript \(T\) indicates matrix transposition. As depth increases, the heatmap becomes progressively lighter, reflecting enhanced orthogonality among prototypes and correspondingly stricter optimization objectives. This shows that DAP effectively adapts to the increasing optimization capacity of deeper layers, providing more suitable objectives for loss computation and performance evaluation at each depth.


\subsection{Discussion on the Parameter-Adaptive Modulation}
\begin{figure}[t]
    \centering
    \subfloat[]{
        \includegraphics[width=0.47\columnwidth]{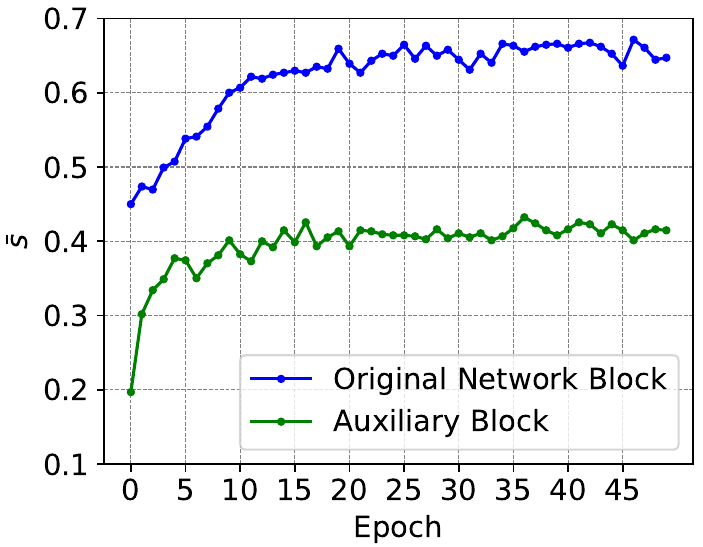}
        \label{fig:decouple_a}
    }
    \hfill
    \subfloat[]{
        \includegraphics[width=0.47\columnwidth]{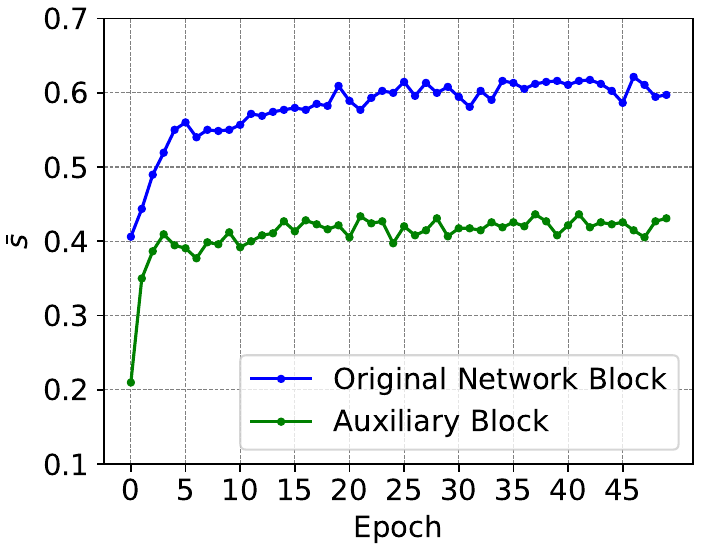}
        \label{fig:decouple_b}
    }
    \caption{Variation of the average performance $\bar{s}$ of visual modality’s original network block and auxiliary block during training on CREMA-D. (a) Both original network block and Auxiliary Block participate in joint training; (b) Only the Auxiliary Block participates in joint training.}
    \label{fig:decouple}
\end{figure}

In this section, we further discuss the parameter decoupling module and the auxiliary-weak interaction strategy in parameter-adaptive modulation. Further experiments are conducted to investigate the performance variation of the original network block and auxiliary block of a modality during AIM training. We report the average performance across depths of a certain modality, which can be calculated as $\bar{s}^m=\frac{1}{D}\sum_{d=1}^D{s_d^m}$. The performance $s_d^m$ at depth $d$ is evaluated using Eq. (\ref{eq:smd}). As shown in Fig. \ref{fig:decouple_a}, for the visual modality on the CREMA-D dataset, the accuracy of the auxiliary block during training consistently remains lower than that of the original network block. This demonstrates that the proposed parameter decoupling mechanism can effectively extract the under-optimized, lower-performing parameters from the original network block. Moreover, the accuracy of the original network block exhibits a growth trend largely consistent with that of the auxiliary block, indicating that it is enhanced through updates in the auxiliary block. The slower accuracy improvement of the original network block compared to the auxiliary block arises from the fact that the auxiliary block contains only a subset of the parameters of the network block.


\begin{figure}[t]
    \centering
    \subfloat[Multimodal Joint-training Baseline]{
        \includegraphics[width=0.47\columnwidth]{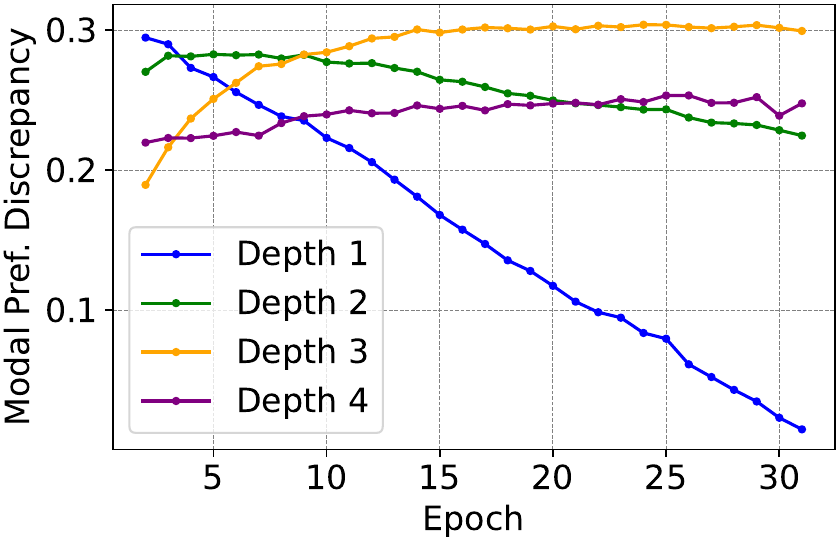}
        \label{fig:observe_a}
    }
    \hfill
    \subfloat[Ours]{
        \includegraphics[width=0.47\columnwidth]{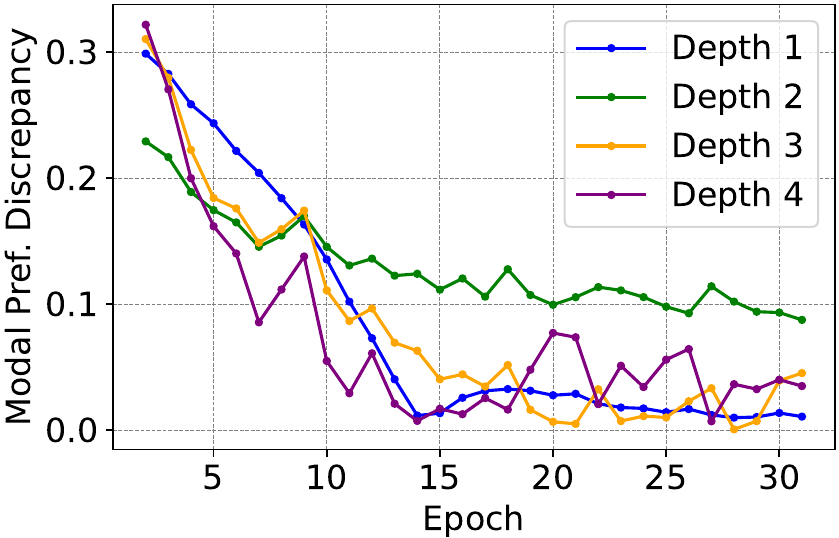}
        \label{fig:observe_b}
    }
    \caption{The variation of modality performance discrepancy at different depths during training on the CREMA-D dataset in (a) the multimodal joint-training baseline and (b) our method. The modality performance discrepancy is calculated via Eq. (\ref{eq:alpha-d}).}
    \label{fig:observe}
\end{figure}

To further verify that the original network block can be enhanced through updates from its auxiliary block, we remove the original network block in the auxiliary-based interaction of PAM. Specifically, we set the loss weight of the original network blocks in Eq. (\ref{eq:l-d}) to zero. The training results are shown in Fig. \ref{fig:decouple_b}. Compared with Fig. \ref{fig:decouple_a}, the accuracy of the original network block in Fig. \ref{fig:decouple_b} does not show a significant drop, indicating that it can be effectively trained solely with its corresponding auxiliary block. Moreover, in Fig. \ref{fig:decouple_b}, the rising phase of the original network block's accuracy is shorter than that in Fig. \ref{fig:decouple_a} and becomes more aligned with the auxiliary block. This is because, in Fig. \ref{fig:decouple_a}, the learning of the original network block also partially relies on its own loss term.

\subsection{Discussion on the Depth-Adaptive Modulation}
To validate the effectiveness of Depth-Adaptive Modulation (DAM), we compare the variation of modality performance discrepancy during training calculated via Eq. (\ref{eq:alpha-d}) across different depths before and after applying our method. As shown in Fig. \ref{fig:observe_a}, in the baseline without our method, the discrepancies at most depths fail to converge to a low level, reflecting substantial modality imbalance. In contrast, Fig. \ref{fig:observe_b} shows that with DAM, the imbalance levels across depths converge to lower values. This indicates that DAM effectively reduces modality imbalance at each depth, thereby enhancing overall balanced multimodal learning. \textbf{More discussion of DAM is provided in the Supplemental Materials.}

\begin{figure}[t]
\centering
\includegraphics[width=\columnwidth]{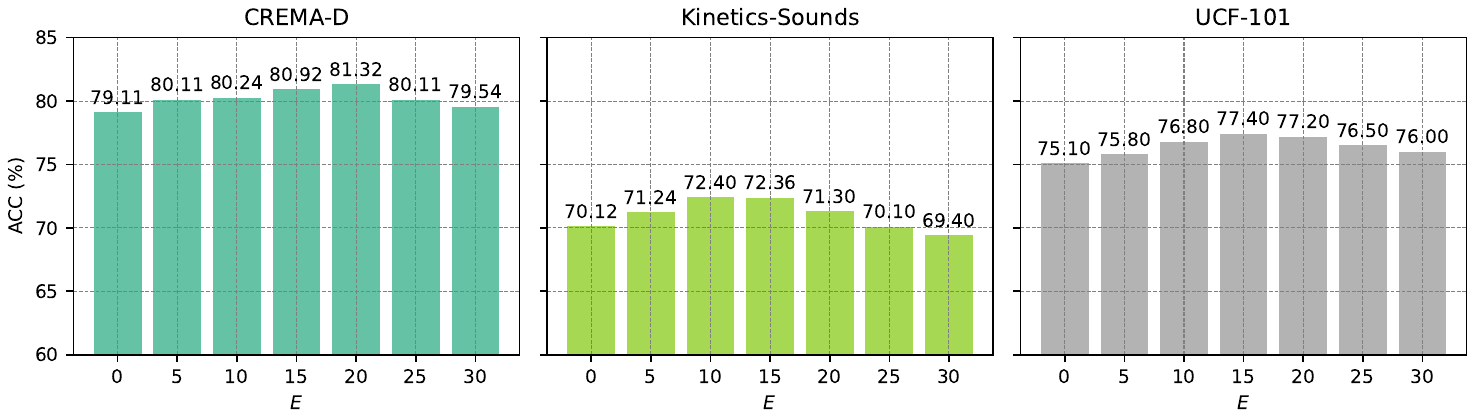} 
\caption{Parameter analysis of $E$. On different datasets, the multimodal accuracy varies smoothly with respect to $E$, exhibiting a consistent trend of first increasing and then decreasing.}
\label{fig:diff-epoch}
\end{figure}

\subsection{Parameter Analysis} \label{sec:e-analysis}
In the multimodal joint training framework incorporating AIM, there is a hyperparameter $E$ indicating the starting epoch of modulation, as mentioned in Algorithm \ref{alg:aim}. Additional experiments are conducted to investigate the impact of $E$ on the training results. Fig. \ref{fig:diff-epoch} presents the multimodal performance for different values of $E$, where $E=0$ indicates that AIM is applied from the beginning of training. As $E$ increases, the accuracy of the model first shows a slight improvement and then declines, remaining relatively stable as $E$ varies. This trend can be explained as follows: when AIM is introduced too early (small $E$), the depth-adaptive prototypes and parameter decoupling mechanism are not yet well trained, leading to less effective modulation in the early stages of training. Conversely, when AIM is introduced too late (large $E$), modality imbalance within the multimodal framework is not addressed promptly, resulting in suboptimal training performance.

\begin{figure}[t]
\centering
\includegraphics[width=0.9\columnwidth]{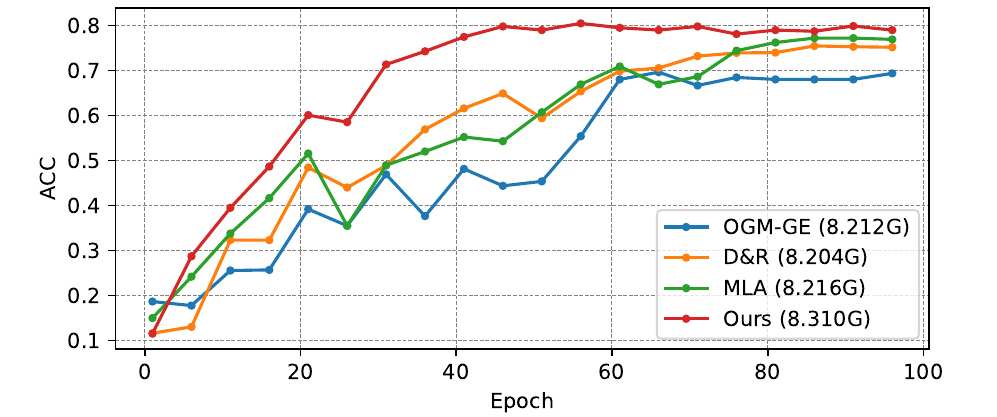} 
\caption{Performance variation of different methods on the CREMA-D dataset using ResNet18 as the encoder backbone. Numbers in parentheses denote training FLOPs per sample.}
\label{fig:train-eff}
\end{figure}

\subsection{Training Efficiency} \label{sec:train-eff}
To evaluate the training efficiency of AIM, we compare its convergence speed and FLOPs with those of SOTA methods for balanced multimodal learning. All methods are trained under identical optimizer settings, data loading procedures, and hardware conditions, with balancing strategies introduced from the beginning of training (Epoch=0). Using ResNet18 as the backbone, AIM exhibits comparable training FLOPs to other methods while converging in fewer epochs. As shown in Fig. \ref{fig:train-eff}, AIM reaches its peak and stable accuracy around the 50th epoch, whereas other methods typically require about 70 epochs. This demonstrates that AIM effectively accelerates overall model learning by fully stimulating the training of slower-optimizing parameters and layers within each modality.

\section{Conclusions}

This paper highlights a commonly overlooked issue in prior works, namely the varying optimization pace across parameters and layers within networks. To address this, we propose Adaptive Intra-Network Modulation (AIM), a training framework comprising parameter-adaptive modulation (PAM) and depth-adaptive modulation (DAM). By adaptively stimulating slow-optimizing parameters and layers, AIM achieves balanced multimodal learning while fully exploiting each modality network's learning potential. In addition, depth-adaptive prototypes facilitate AIM by providing an effective method to estimate performance across network components. Extensive experiments demonstrate AIM's superiority over SOTA balanced multimodal learning methods, the contributions of its components, and its effectiveness across various backbones, fusion strategies, and optimizers.

While current balanced multimodal learning methods perform well on classification, their effectiveness on regression remains underexplored. In future work, we will further extend AIM to a broader range of tasks, including regression.

\bibliographystyle{IEEEtran}
\bibliography{reference}

\begin{thebibliography}{10}
\providecommand{\url}[1]{#1}
\csname url@samestyle\endcsname
\providecommand{\newblock}{\relax}
\providecommand{\bibinfo}[2]{#2}
\providecommand{\BIBentrySTDinterwordspacing}{\spaceskip=0pt\relax}
\providecommand{\BIBentryALTinterwordstretchfactor}{4}
\providecommand{\BIBentryALTinterwordspacing}{\spaceskip=\fontdimen2\font plus
\BIBentryALTinterwordstretchfactor\fontdimen3\font minus \fontdimen4\font\relax}
\providecommand{\BIBforeignlanguage}[2]{{%
\expandafter\ifx\csname l@#1\endcsname\relax
\typeout{** WARNING: IEEEtran.bst: No hyphenation pattern has been}%
\typeout{** loaded for the language `#1'. Using the pattern for}%
\typeout{** the default language instead.}%
\else
\language=\csname l@#1\endcsname
\fi
#2}}
\providecommand{\BIBdecl}{\relax}
\BIBdecl

\bibitem{gazzaniga2006cognitive}
M.~S. Gazzaniga, R.~B. Ivry, and G.~Mangun, ``Cognitive neuroscience. the biology of the mind,(2014),'' 2006.

\bibitem{herreras2010cognitive}
E.~B. Herreras, ``Cognitive neuroscience; the biology of the mind,'' \emph{Cuadernos de Neuropsicolog{\'\i}a/Panamerican Journal of Neuropsychology}, vol.~4, no.~1, pp. 87--90, 2010.

\bibitem{dou2022empirical}
Z.-Y. Dou, Y.~Xu, Z.~Gan, J.~Wang, S.~Wang, L.~Wang, C.~Zhu, P.~Zhang, L.~Yuan, N.~Peng \emph{et~al.}, ``An empirical study of training end-to-end vision-and-language transformers,'' in \emph{Proceedings of the IEEE/CVF Conference on Computer Vision and Pattern Recognition}, 2022, pp. 18\,166--18\,176.

\bibitem{shankar2018review}
S.~K. Shankar, L.~P. Prieto, M.~J. Rodr{\'\i}guez-Triana, and A.~Ruiz-Calleja, ``A review of multimodal learning analytics architectures,'' in \emph{2018 IEEE 18th international conference on advanced learning technologies (ICALT)}.\hskip 1em plus 0.5em minus 0.4em\relax IEEE, 2018, pp. 212--214.

\bibitem{10339893}
R.~Huan, G.~Zhong, P.~Chen, and R.~Liang, ``Unimf: A unified multimodal framework for multimodal sentiment analysis in missing modalities and unaligned multimodal sequences,'' \emph{IEEE Transactions on Multimedia}, vol.~26, pp. 5753--5768, 2024.

\bibitem{liang2022foundations}
P.~P. Liang, A.~Zadeh, and L.-P. Morency, ``Foundations and trends in multimodal machine learning: Principles, challenges, and open questions,'' \emph{arXiv preprint arXiv:2209.03430}, 2022.

\bibitem{baltruvsaitis2018multimodal}
T.~Baltru{\v{s}}aitis, C.~Ahuja, and L.-P. Morency, ``Multimodal machine learning: A survey and taxonomy,'' \emph{IEEE transactions on pattern analysis and machine intelligence}, vol.~41, no.~2, pp. 423--443, 2018.

\bibitem{10814984}
T.-Y. Kim, J.~Yang, and E.~Park, ``Msdlf-k: A multimodal feature learning approach for sentiment analysis in korean incorporating text and speech,'' \emph{IEEE Transactions on Multimedia}, vol.~27, pp. 1266--1276, 2025.

\bibitem{9681296}
X.~Guo, A.~W.-K. Kong, and A.~Kot, ``Deep multimodal sequence fusion by regularized expressive representation distillation,'' \emph{IEEE Transactions on Multimedia}, vol.~25, pp. 2085--2096, 2023.

\bibitem{10171388}
D.~Chen and R.~Zhang, ``Building multimodal knowledge bases with multimodal computational sequences and generative adversarial networks,'' \emph{IEEE Transactions on Multimedia}, vol.~26, pp. 2027--2040, 2024.

\bibitem{kazakos2019epic}
E.~Kazakos, A.~Nagrani, A.~Zisserman, and D.~Damen, ``Epic-fusion: Audio-visual temporal binding for egocentric action recognition,'' in \emph{Proceedings of the IEEE/CVF international conference on computer vision}, 2019, pp. 5492--5501.

\bibitem{gao2020listen}
R.~Gao, T.-H. Oh, K.~Grauman, and L.~Torresani, ``Listen to look: Action recognition by previewing audio,'' in \emph{Proceedings of the IEEE/CVF conference on computer vision and pattern recognition}, 2020, pp. 10\,457--10\,467.

\bibitem{hazarika2020misa}
D.~Hazarika, R.~Zimmermann, and S.~Poria, ``Misa: Modality-invariant and-specific representations for multimodal sentiment analysis,'' in \emph{Proceedings of the 28th ACM international conference on multimedia}, 2020, pp. 1122--1131.

\bibitem{sun2022cubemlp}
H.~Sun, H.~Wang, J.~Liu, Y.-W. Chen, and L.~Lin, ``Cubemlp: An mlp-based model for multimodal sentiment analysis and depression estimation,'' in \emph{Proceedings of the 30th ACM international conference on multimedia}, 2022, pp. 3722--3729.

\bibitem{potamianos2004audio}
G.~Potamianos, C.~Neti, J.~Luettin, and I.~Matthews, ``Audio-visual automatic speech recognition: An overview,'' \emph{Issues in visual and audio-visual speech processing}, vol.~22, p.~23, 2004.

\bibitem{8387512}
H.~Li, J.~Zhu, C.~Ma, J.~Zhang, and C.~Zong, ``Read, watch, listen, and summarize: Multi-modal summarization for asynchronous text, image, audio and video,'' \emph{IEEE Transactions on Knowledge and Data Engineering}, vol.~31, no.~5, pp. 996--1009, 2019.

\bibitem{10445009}
H.~Liu, C.~L.~P. Chen, X.~Gong, and T.~Zhang, ``Robust saliency-aware distillation for few-shot fine-grained visual recognition,'' \emph{IEEE Transactions on Multimedia}, vol.~26, pp. 7529--7542, 2024.

\bibitem{1468162}
G.~Ghinea and J.~Thomas, ``Quality of perception: user quality of service in multimedia presentations,'' \emph{IEEE Transactions on Multimedia}, vol.~7, no.~4, pp. 786--789, 2005.

\bibitem{peng2022balanced}
X.~Peng, Y.~Wei, A.~Deng, D.~Wang, and D.~Hu, ``Balanced multimodal learning via on-the-fly gradient modulation,'' in \emph{Proceedings of the IEEE/CVF conference on computer vision and pattern recognition}, 2022, pp. 8238--8247.

\bibitem{fan2023pmr}
Y.~Fan, W.~Xu, H.~Wang, J.~Wang, and S.~Guo, ``Pmr: Prototypical modal rebalance for multimodal learning,'' in \emph{Proceedings of the IEEE/CVF Conference on Computer Vision and Pattern Recognition}, 2023, pp. 20\,029--20\,038.

\bibitem{zhang2024multimodal}
X.~Zhang, J.~Yoon, M.~Bansal, and H.~Yao, ``Multimodal representation learning by alternating unimodal adaptation,'' in \emph{Proceedings of the IEEE/CVF conference on computer vision and pattern recognition}, 2024, pp. 27\,456--27\,466.

\bibitem{wei2024diagnosing}
Y.~Wei, S.~Li, R.~Feng, and D.~Hu, ``Diagnosing and re-learning for balanced multimodal learning,'' in \emph{European Conference on Computer Vision}.\hskip 1em plus 0.5em minus 0.4em\relax Springer, 2024, pp. 71--86.

\bibitem{huang2022modality}
Y.~Huang, J.~Lin, C.~Zhou, H.~Yang, and L.~Huang, ``Modality competition: What makes joint training of multi-modal network fail in deep learning?(provably),'' in \emph{International conference on machine learning}.\hskip 1em plus 0.5em minus 0.4em\relax PMLR, 2022, pp. 9226--9259.

\bibitem{wang2020makes}
W.~Wang, D.~Tran, and M.~Feiszli, ``What makes training multi-modal classification networks hard?'' in \emph{Proceedings of the IEEE/CVF conference on computer vision and pattern recognition}, 2020, pp. 12\,695--12\,705.

\bibitem{wei2024fly}
Y.~Wei, D.~Hu, H.~Du, and J.-R. Wen, ``On-the-fly modulation for balanced multimodal learning,'' \emph{IEEE Transactions on Pattern Analysis and Machine Intelligence}, 2024.

\bibitem{wu2022characterizing}
N.~Wu, S.~Jastrzebski, K.~Cho, and K.~J. Geras, ``Characterizing and overcoming the greedy nature of learning in multi-modal deep neural networks,'' in \emph{International Conference on Machine Learning}.\hskip 1em plus 0.5em minus 0.4em\relax PMLR, 2022, pp. 24\,043--24\,055.

\bibitem{saratchandran2024activation}
H.~Saratchandran, S.~Ramasinghe, and S.~Lucey, ``From activation to initialization: Scaling insights for optimizing neural fields,'' in \emph{Proceedings of the IEEE/CVF Conference on Computer Vision and Pattern Recognition}, 2024, pp. 413--422.

\bibitem{chen2023layer}
Y.~Chen, A.~Yuille, and Z.~Zhou, ``Which layer is learning faster? a systematic exploration of layer-wise convergence rate for deep neural networks,'' in \emph{The Eleventh International Conference on Learning Representations}, 2023.

\bibitem{liu2025optimization}
L.~Liu, X.~Cao, H.~Wang, and J.~Xiang, ``Optimization of model parameters and hyperparameters in deep learning models for spatial interaction prediction,'' \emph{Expert Systems with Applications}, vol. 266, p. 126160, 2025.

\bibitem{ramachandram2017deep}
D.~Ramachandram and G.~W. Taylor, ``Deep multimodal learning: A survey on recent advances and trends,'' \emph{IEEE signal processing magazine}, vol.~34, no.~6, pp. 96--108, 2017.

\bibitem{zhu2024vision+}
Y.~Zhu, Y.~Wu, N.~Sebe, and Y.~Yan, ``Vision+ x: A survey on multimodal learning in the light of data,'' \emph{IEEE Transactions on Pattern Analysis and Machine Intelligence}, 2024.

\bibitem{10076804}
S.~Li, T.~Zhang, B.~Chen, and C.~L.~P. Chen, ``Mia-net: Multi-modal interactive attention network for multi-modal affective analysis,'' \emph{IEEE Transactions on Affective Computing}, vol.~14, no.~4, pp. 2796--2809, 2023.

\bibitem{9863920}
T.~Zhang, S.~Li, B.~Chen, H.~Yuan, and C.~L. Philip~Chen, ``Aia-net: Adaptive interactive attention network for text–audio emotion recognition,'' \emph{IEEE Transactions on Cybernetics}, vol.~53, no.~12, pp. 7659--7671, Dec 2023.

\bibitem{10577436}
S.~Li, T.~Zhang, and C.~L.~P. Chen, ``Sia-net: Sparse interactive attention network for multimodal emotion recognition,'' \emph{IEEE Transactions on Computational Social Systems}, vol.~11, no.~5, pp. 6782--6794, 2024.

\bibitem{11049910}
------, ``Cyclic data distillation semi-supervised learning for multi-modal emotion recognition,'' \emph{IEEE Transactions on Knowledge and Data Engineering}, pp. 1--14, 2025.

\bibitem{zhang2023provable}
Q.~Zhang, H.~Wu, C.~Zhang, Q.~Hu, H.~Fu, J.~T. Zhou, and X.~Peng, ``Provable dynamic fusion for low-quality multimodal data,'' in \emph{International conference on machine learning}.\hskip 1em plus 0.5em minus 0.4em\relax PMLR, 2023, pp. 41\,753--41\,769.

\bibitem{9767641}
S.~Mai, Y.~Zeng, and H.~Hu, ``Multimodal information bottleneck: Learning minimal sufficient unimodal and multimodal representations,'' \emph{IEEE Transactions on Multimedia}, vol.~25, pp. 4121--4134, 2023.

\bibitem{10269037}
Z.~Ding, G.~Lan, Y.~Song, and Z.~Yang, ``Sgir: Star graph-based interaction for efficient and robust multimodal representation,'' \emph{IEEE Transactions on Multimedia}, vol.~26, pp. 4217--4229, 2024.

\bibitem{10224356}
S.~Mai, Y.~Sun, A.~Xiong, Y.~Zeng, and H.~Hu, ``Multimodal boosting: Addressing noisy modalities and identifying modality contribution,'' \emph{IEEE Transactions on Multimedia}, vol.~26, pp. 3018--3033, 2024.

\bibitem{10261246}
L.~Wei, D.~Hu, W.~Zhou, and S.~Hu, ``Modeling both intra- and inter-modality uncertainty for multimodal fake news detection,'' \emph{IEEE Transactions on Multimedia}, vol.~25, pp. 7906--7916, 2023.

\bibitem{sun2021learning}
Y.~Sun, S.~Mai, and H.~Hu, ``Learning to balance the learning rates between various modalities via adaptive tracking factor,'' \emph{IEEE Signal Processing Letters}, vol.~28, pp. 1650--1654, 2021.

\bibitem{abdi2010coefficient}
H.~Abdi, ``Coefficient of variation,'' \emph{Encyclopedia of research design}, vol.~1, no.~5, pp. 169--171, 2010.

\bibitem{cao2014crema}
H.~Cao, D.~G. Cooper, M.~K. Keutmann, R.~C. Gur, A.~Nenkova, and R.~Verma, ``Crema-d: Crowd-sourced emotional multimodal actors dataset,'' \emph{IEEE transactions on affective computing}, vol.~5, no.~4, pp. 377--390, 2014.

\bibitem{arandjelovic2017look}
R.~Arandjelovic and A.~Zisserman, ``Look, listen and learn,'' in \emph{Proceedings of the IEEE international conference on computer vision}, 2017, pp. 609--617.

\bibitem{kay2017kinetics}
W.~Kay, J.~Carreira, K.~Simonyan, B.~Zhang, C.~Hillier, S.~Vijayanarasimhan, F.~Viola, T.~Green, T.~Back, P.~Natsev \emph{et~al.}, ``The kinetics human action video dataset,'' \emph{arXiv preprint arXiv:1705.06950}, 2017.

\bibitem{soomro2012ucf101}
K.~Soomro, A.~R. Zamir, and M.~Shah, ``Ucf101: A dataset of 101 human actions classes from videos in the wild,'' \emph{arXiv preprint arXiv:1212.0402}, 2012.

\bibitem{zadeh2016mosi}
A.~Zadeh, R.~Zellers, E.~Pincus, and L.-P. Morency, ``Mosi: multimodal corpus of sentiment intensity and subjectivity analysis in online opinion videos,'' \emph{arXiv preprint arXiv:1606.06259}, 2016.

\bibitem{li2023boosting}
H.~Li, X.~Li, P.~Hu, Y.~Lei, C.~Li, and Y.~Zhou, ``Boosting multi-modal model performance with adaptive gradient modulation,'' in \emph{Proceedings of the IEEE/CVF International Conference on Computer Vision}, 2023, pp. 22\,214--22\,224.

\bibitem{liang2021multibench}
P.~P. Liang, Y.~Lyu, X.~Fan, Z.~Wu, Y.~Cheng, J.~Wu, L.~Chen, P.~Wu, M.~A. Lee, Y.~Zhu \emph{et~al.}, ``Multibench: Multiscale benchmarks for multimodal representation learning,'' \emph{Advances in neural information processing systems}, vol. 2021, no. DB1, p.~1, 2021.

\end{thebibliography}

\end{document}